\documentclass[letterpaper]{article} 
\usepackage{aaai24}  
\usepackage{times}  
\usepackage{helvet}  
\usepackage{courier}  
\usepackage[hyphens]{url}  
\usepackage{graphicx} 
\usepackage{natbib}  
\usepackage{caption} 
\usepackage{algorithm}

\usepackage{newfloat}
\usepackage{listings}
\DeclareCaptionStyle{ruled}{labelfont=normalfont,labelsep=colon,strut=off} 
\floatstyle{ruled}
\newfloat{listing}{tb}{lst}{}
\floatname{listing}{Listing}
\usepackage{url}            
\usepackage{booktabs}       
\usepackage{amsfonts}       
\usepackage{bm}
\usepackage{algpseudocode}
\usepackage{enumitem}
\usepackage{xcolor}

\usepackage{subfigure}

\usepackage{amsmath}
\usepackage{amssymb}
\usepackage{mathtools}
\usepackage{amsthm}
\DeclareMathAlphabet \mathbfcal{OMS}{cmsy}{b}{n}

\newcommand{\ten}[1]{\mathbfcal{#1}} 
\newcommand{\mat}[1]{\mathbf{#1}}

\nocopyright 

\title{Tensor-Compressed Back-Propagation-Free Training for\\ (Physics-Informed) Neural Networks}

\author {
    Yequan Zhao\textsuperscript{\rm 1, \rm a},
    Xinling Yu\textsuperscript{\rm 1, \rm a},
    Zhixiong Chen\textsuperscript{\rm 1},
    Ziyue Liu\textsuperscript{\rm 1},
    Sijia Liu\textsuperscript{\rm 2, \rm 3},
    Zheng Zhang\textsuperscript{\rm 1}
}
\affiliations {
    \textsuperscript{\rm 1} University of California, Santa Barbara, CA 93106\\
    \textsuperscript{\rm 2} Michigan State University, \textsuperscript{\rm 3} MIT-IBM Watson AI Lab\\
    \{yequan\_zhao, xyu644, zhixiong, ziyueliu, zzhang01\}@ucsb.edu, liusiji5@msu.edu \\
    \textsuperscript{\rm a} Equal contributions.
}

\usepackage{bibentry}

\begin{document}

\maketitle

\begin{abstract}
Backward propagation (BP) is widely used to compute the gradients in neural network training. However, it is hard to implement BP on edge devices due to the lack of hardware and software resources to support automatic differentiation. This has tremendously increased the design complexity and time-to-market of on-device training accelerators. This paper presents a completely BP-free framework that only requires forward propagation to train realistic neural networks. Our technical contributions are three-fold. Firstly, we present a tensor-compressed variance reduction approach to greatly improve the scalability of zeroth-order (ZO) optimization, making it feasible to handle a network size that is beyond the capability of previous ZO approaches. Secondly, we present a hybrid gradient evaluation approach to improve the efficiency of ZO training. Finally, we extend our BP-free training framework to physics-informed neural networks (PINNs) by proposing a sparse-grid approach to estimate the derivatives in the loss function without using BP. Our BP-free training only loses little accuracy on the MNIST dataset compared with standard  first-order training. We also demonstrate successful results in training a PINN for solving a 20-dim Hamiltonian-Jacobi-Bellman PDE. This memory-efficient and BP-free approach may serve as a foundation for the near-future on-device training on many resource-constraint platforms (e.g., FPGA, ASIC, micro-controllers, and photonic chips). 
\end{abstract}

\section{Introduction}

 In neural network training, it is widely assumed that the gradient information can be computed via backward propagation (BP)~\cite{rumelhart1986learning} to perform SGD-type optimization. This is true on HPC and desktop computers, because they support Pytorch/Tensorflow backend and automatic differentiation (AD) packages~\cite{bolte2020mathematical} to compute exact gradients. However, performing BP is infeasible on many edge devices (e.g., FPGA, ASIC, photonic chips, embedded micro-processors) due to the lack of necessary computing and memory resources to support AD libraries.  Manually calculating the gradient of a modern neural network on edge hardware is time-consuming due to the high complexity and lots of debug iterations. This can significantly delay the development and deployment of AI training accelerators. For instance, designing an FPGA inference accelerator may be done within one week via high-level synthesis by an experienced AI hardware expert, but designing an FPGA training accelerator can take one or two years due to the high complexity of implementing BP. Some analog edge computing platforms (e.g., photonic chips) even do not have scalable memory arrays to store the final or intermediate results of a chain rule-based gradient computation. 

This paper investigates the end-to-end training of neural networks and physics-informed neural networks (PINNs) without using BP, which can potentially make on-device neural network training accelerator design as easy as designing an inference accelerator. The demand for edge-device training has been growing rapidly in recent years. One motivation is to ensure AI model performance under varying data stream and environment, as well as device parameter shift. Another motivation is the increasing concerns about data privacy, which requires end-to-end or incremental learning on edge devices based on local data~\cite{mcmahan2017communication}. In science and engineering, PINNs~\cite{raissi2019physics,lagaris1998artificial} have been increasingly used to solve the forward and inverse problems of high-dimensional partial differential equations (PDE). A PINN needs to be trained again once the PDE initial conditions, boundary conditions, or measurement data changes. Applications of such on-device PINN training include (but are not limited to) safety-aware verification and control of autonomous systems~\cite{bansal2021deepreach,onken2020neural,sun2020learning} and MRI-based electrical property tomography~\cite{yu2023pifon}.

We intend to perform BP-free training via stochastic zeroth-order (ZO) optimization~
\cite{duchi2015optimal,nesterov2017random,ZOSGD,shamir2017optimal,balasubramanian2022zeroth,liu2020primer}, 
which uses a few forward evaluations with perturbed model parameters to approximate the true gradient. In the realm of deep learning, ZO optimization was primarily used for crafting black-box adversarial examples to assess neural network robustness~\cite{chen2017zoo, liu2020min}, and for parameter and memory-efficient model fine-tuning~\cite{malladi2023fine,zhang2022how}. Notably, ZO optimization has been rarely used in neural network training from scratch, because the variance of the ZO gradient estimation is large when the number of training variables increases. 

Leveraging ZO optimization, we present a scalable BP-free approach for training both standard neural networks and PINNs from scratch. Our novel contributions include:
\begin{itemize}[leftmargin=*]
    \item{\bf A hybrid and tensor-compressed ZO training approach.} We first reduce the gradient variance of ZO optimization significantly via a tensor-compressed  dimension reduction in the training process. This variance reduction approach enables training realistic neural networks and PINNs which are beyond the capability of existing ZO training. Then we design a hybrid approach that combines the advantages of random ZO gradient estimation and finite-difference methods to reduce the number of forward evaluations in ZO training. 
    \item{\bf A completely BP-free approach for training PINNs.}  The loss function of PINNs often involves differential operators, which complicates the loss evaluation and the training process on resource-constraint edge devices. To avoid using BP in the loss evaluation, we propose a sparse-grid method to implement the Stein estimator. This allows solving vast (high-dimensional) PDE problems on edge devices where the BP computing engines are not available. 
    \item {\bf Numerical experiments.} We validate our method on the MNIST image classification task and in solving a high-dim PDE. We show that the tensor-compressed method can greatly improve the scalability of ZO training, showing better performance than sparsity-based ZO training. Our BP-free approach also shows high accuracy in solving a 20-dim Hamiltonian-Jacobi-Bellman (HJB) PDE, which is a critical task in optimal control.
\end{itemize} 
While our approach still needs improvement in order to train very large neural networks (e.g., transformers), it is good enough to enable on-device training for many image, speech and scientific computing problems. 


\section{Background}
This section introduces the necessary background of ZO optimization and PINN. 

 \subsection{Zeroth-Order (ZO) Optimization}
        
        We consider the minimization of a loss function $\mathcal{L}(\bm{\theta})$ by updating the model parameters $\bm{\theta} \in \mathbb{R}^d$ iteratively using a (stochastic) gradient descent method: 
        \begin{equation}
            \bm{\theta}_t \leftarrow \bm{\theta}_{t-1}-\alpha \bm{g}
            \label{SGD update}
        \end{equation}
        where $\bm{g}$ denotes the (stochastic) gradient of the loss $\mathcal{L}$ w.r.t. model parameters $\bm{\theta}$. In some cases it is hard or even impossible to exactly compute the gradient vector $\bm{\theta}$, therefore we consider ZO optimization, which uses some forward function queries to approximate the gradient $\bm{g}$:
        \begin{equation}
           \bm{g} \approx \hat{\nabla}_{\bm{\theta}}\mathcal{L}(\bm{\theta})=
            \sum_{i=1}^N \frac{1}{N\mu} \left[\mathcal{L}\left(\bm{\theta}+\mu \bm{\xi}_i\right)-\mathcal{L}(\bm{\theta})\right] \bm{\xi}_i.
        \label{ZO gradient estimation}
        \end{equation}
        Here $\{\bm{\xi}_i\}_{i=1}^N$ are some perturbation vectors and $\mu$ is the sampling radius, which is typically small. Two perturbation methods are commonly used:
        \begin{itemize}[leftmargin=*]
            \item {\bf Random gradient estimators (RGE)}:  $\{\bm{\xi}_i\}_{i=1}^N$ are $N$ i.i.d. samples drawn from a distribution $\rho(\bm{\xi})$ (e.g., a multivariate Gaussian distribution). 
            \item {\bf Coordinate-wise gradient estimator (CGE)} [or finite difference (FD)]: $N=d$ and $\bm{\xi}_i$ denotes the \textit{i}-th elementary basis vector, with one at the \textit{i}-th coordinate and zeros elsewhere.
        \end{itemize}

        In both cases, the expectation of $\hat{\nabla}_{\bm{\theta}}\mathcal{L}$ is unbiased w.r.t. the gradient of the smoothed function $f_\mu(\bm{x}):=\mathbb{E}_{\bm{\xi} \sim \rho(\bm{\xi})}[f(\bm{x}+\mu \bm{\xi})]$, however biased w.r.t. the true gradient $\nabla_{\bm{\theta}}\mathcal{L}$~\cite{berahas2022theoretical}. 
        In comparison to CGE, RGE is typically more query-efficient, requiring a less number of function evaluations $N$ compared to the number of optimization variables $d$. 
        The variances of both RGE and CGE involve a dimension-dependent factor $O(d/N)$ given $\mu = O(1/\sqrt{N})$~\cite{liu2020primer}.

        Since the MSE error increases quickly as $d$ increases, ZO optimization was mainly used in some small- or medium-size 
        optimization problems, such as black-box adversarial attack generation and memory-efficient fine-tuning.

 \subsection{Physics-Informed Neural Networks (PINNs)}
Consider a generic partial differential equation (PDE):
\begin{equation}
\begin{aligned}
\mathcal{N}[\bm{u}(\bm{x},t)]&=l(\bm{x}, t), ~\quad \bm{x} \in \Omega,~~t \in[0, T],\\
\mathcal{I}[\bm{u}(\bm{x},0)] &= g(\bm{x}), ~~~\quad \bm{x} \in \Omega,\\
\mathcal{B}[\bm{u}(\bm{x},t)] &= h(\bm{x}, t),  \quad \boldsymbol{x} \in \partial \Omega, ~~t \in[0, T],  
\end{aligned}
\label{general PDE}
\end{equation}
where $\bm{x}$ and $t$ are the spatial and temporal coordinates; $\Omega \subset \mathbb{R}^{D}$, $\partial \Omega$ and $T$ denote the spatial domain, domain boundary and time horizon, respectively; $\mathcal{N}$ is a general nonlinear differential operator; $\mathcal{I}$ and $\mathcal{B}$ represent the initial (or terminal) and boundary condition; $\bm{u} \in \mathbb{R}^{n}$ is the solution for the PDE described above. In the contexts of PINNs~\cite{raissi2019physics}, a solution network $\bm{u}_{\bm{\theta}}(\bm{x},t)$, parameterized by $\bm{\theta}$, is substituted into PDE \eqref{general PDE}, resulting in a residual defined as:
\begin{equation}
r_{\bm{\theta}}(\bm{x},t):=\mathcal{N}[\bm{u}_{\bm{\theta}}(\bm{x},t)]-l(\bm{x}, t).
\label{PDE residual}
\end{equation}
The parameters $\bm{\theta}$ can be trained by minimizing the loss:
\begin{equation}
\mathcal{L}(\bm{\theta})=\mathcal{L}_r(\bm{\theta})+\lambda_{0}\mathcal{L}_0(\bm{\theta})+\lambda_{b}\mathcal{L}_b (\bm{\theta}).
\label{PINNs loss}
\end{equation}
Here
\begin{equation}
\begin{aligned}
\mathcal{L}_r (\bm{\theta}) &= \frac{1}{N_{r}}\sum_{i=1}^{N_{r}}  \left\|r_{\bm{\theta}}(\bm{x_{r}}^{i},t_{r}^{i})\right\|_{2}^{2},\\
\mathcal{L}_0 (\bm{\theta}) &= \frac{1}{N_{0}}\sum_{i=1}^{N_{0}}\left\|\mathcal{I}[\bm{u}_{\bm{\theta}}(\bm{x_{0}}^{i},0)] - g(\bm{x_{0}}^{i})\right\|_{2}^{2},\\
\text{and}\quad\mathcal{L}_b (\bm{\theta}) &= \frac{1}{N_{b}}\sum_{i=1}^{N_{b}}\left\|\mathcal{B}[\bm{u}_{\bm{\theta}}(\bm{x_{b}}^{i},t_{b}^{i})] - h(\bm{x_{b}}^{i}, t_{b}^{i})\right\|_{2}^{2}
\end{aligned}
\label{loss terms}
\end{equation}
are the residuals of the PDE, the initial (or terminal) condition and boundary condition, respectively. 


\section{Scaling Up ZO Optimization for Neural Network Training}
 
This section improves the ZO method to train real-size neural networks. We first present tensor-compressed approach to dramatically reduce the gradient error and improve the scalability of ZO training. Then we present a hybrid ZO training strategy that achieves a better trade-off between accuracy and function query complexity. 

\vspace{-5pt}
\subsection{Tensor-Train (TT) Variance Reduction}
\label{subsec:TT}

     Our goal is to develop a BP-free training framework based on ZO optimization. As explained before, the ZO gradient estimation error  is dominated by $d$, which is the number of total trainable variables. In realistic neural network training, $d$ can easily exceed $10^5$ even for very simple image classification problems. This prevents the direct application of ZO optimization in end-to-end neural network training. 
      
      To improve the scalability of ZO training, we propose to significantly reduce the dimensionality and gradient MSE error via a \textit{low-rank} tensor-compressed training. Let $\bm{W} \in \mathbb{R}^{M\times N}$ be a generic weight matrix in a neural network. We factorize its dimension sizes as $M = \prod^{L}_{i=1}m_i$ and $N = \prod^{L}_{j=1}n_j$, fold $\bm{W}$ into a $2L$-way tensor $\mathbfcal{W} \in \mathbb{R}^{m_1\times m_2 \times \dots \times m_L \times n_1 \times n_2 \times \dots \times n_L}$, and parameterize $\ten{W}$ with the tensor-train (TT) decomposition \cite{oseledets2011tensor}:
        \begin{equation}
        {\mathbfcal{W}}(i_1, i_2, \dots, i_L, j_1, j_2, \dots, j_L)
        \approx \prod^{L} \limits_{k=1} \mat{G}_k(i_k, j_k)
        \end{equation}
        Here $\mat{G}_k(i_k, j_k) \in \mathbb{R}^{r_{k-1} \times r_{k}}$ is the $(i_k, j_k)$-th slice of the TT-core $\mathbfcal{G}_k \in \mathbb{R}^{r_{k-1}\times m_k \times n_k \times r_k}$by fixing its $2$nd index as $i_k$ and $3$rd index as $j_k$. The vector $(r_0, r_1, \dots, r_{L})$ is called TT-ranks with the constraint $r_0 = r_{L} = 1$. This TT representation reduces the number of unknown variables from $\prod^{L} \limits_{k=1}m_k n_k$ to $\sum \limits_{k=1}^{L}r_{k-1}m_k n_k r_k$. The compression ratio can be controlled by the TT-ranks, which can be learnt automatically via the Bayesian tensor rank determination~\cite{hawkins2021bayesian,hawkins2022towards}. 
        
        In the ZO training process, we change the training variables from $\mat{W}$ to the TT factors $\{\mathbfcal{G}_k \}_{k=1}^L$. This reduces the problem dimensionality $d$ by several orders of magnitude, leading to dramatic reduction of the variance in the RGE gradient estimation and of the query complexity in the CGE gradient estimator. In the ZO training, we only need forward evaluations, where the original matrix-vector product is replaced with low-cost tensor-network contraction. This offers both memory and computing cost reduction in the ZO training process. While we assume $\bm{W}$ as a weight matrix, other model parameters like embedding tables and convolution filters can be compressed simulatenously in the same way (with adjustable ranks) in the ZO training process.

       %

        

    \subsection{A Hybrid ZO Optimizer}
    \label{subsec:hybrid}
With the above tensor-compressed variance reduction, we can employ either RGE or CGE for ZO gradient estimation and perform BP-free training. In practice, the RGE method converges very slowly in the late stage of training due to the large gradient error. CGE needs fewer training epochs due to more accurate gradient estimation, but it needs many more forward evaluations per gradient estimation since it only perturbs one model parameter in each forward evaluation.

        To enhance both the accuracy and efficiency of the whole ZO training process, we employ a hybrid ZO training scheme that involves two stages:
        \begin{itemize}[leftmargin=*]
        \item{\bf ZO-signRGE coarse training:} 
        The RGE method perturbs all model parameters simultaneously with a single random perturbation, requiring significantly fewer forward queries per epoch compared to the CGE method. 
        However, the two-level stochasticity (in SGD and in the gradient estimation, respectively) of ZO via RGE (ZO-RGE) may  cause a large gradient variance and lead to divergence in high-dimensional tasks. To address this issue, we adopt the concept from signSGD~\cite{bernstein2018signsgd} and its ZO counterpart, ZO-signSGD~\cite{liu2019signsgd}, to de-noise the ZO gradient estimation by preserving only the sign for each update:
        \small
     \begin{equation}
               \bm{\theta}_t \leftarrow \bm{\theta}_{t-1}-\alpha {\rm sign}\left[\sum_{i=1}^N \frac{1}{N\mu} \left[\mathcal{L}\left(\bm{\theta}_{t-1}+\mu \bm{\xi}_i\right)-\mathcal{L}(\bm{\theta}_{t-1})\right] \bm{\xi}_i\right].\nonumber
           \end{equation} \normalsize
         We refer to this method as ZO-signRGE. ZO-signRGE retains only the sign of each gradient estimation, mitigating the adverse effects of the high variance of RGE. As a result, it exhibits better robustness to gradient noise and demonstrates faster empirical convergence.

        \item{\bf ZO-CGE fine-tuning:} The CGE method requires $d+1$ forward evaluations per gradient estimation. 
        Therefore, the ZO training with CGE (ZO-CGE) requires many function queries in the whole training process.
        To accelerate training, we further adopt the idea of momentum~\cite{polyak1964some}.  The momentum method accumulates a velocity vector in directions of persistent reduction in the objective across iterations~\cite{sutskever2013importance}. The descent direction $\bm{b}^t$ is given by an exponential moving average of the past gradients $\bm{b}^t \leftarrow m \bm{b}^{t-1} + \bm{g}^t$. Here $m$ is the momentum term, $\bm{g}^t$ is the estimated gradient vector at time $t$, and $\bm{b}^0=\bm{g}^0$.

        \end{itemize}
     The ZO-signRGE coarse training rapidly explores a roughly converged solution with a small number of loss evaluations. 
        When the coarse training fails to learn (e.g., the training loss exhibits trivial updates for several epochs), the optimizer switches to ZO-CGE to fine-tune the model. 

\section{BP-free Training for PINN}

In this section, we extend the proposed TT-compressed hybrid ZO training to PINNs. Training a PINN is more challenging because the loss function in \eqref{PINNs loss} involves first or even high-order derivatives. We intend to also avoid BP computation in the loss evaluation.

\subsection{Stein Gradient Estimation}
Without loss of generality, for an input $\bm{x}\in \mathbb{R}^{D}$ and an approximated PDE solution $\bm{u}_{\bm{\theta}}(\bm{x})\in \mathbb{R}^{n}$ parameterized by $\bm{\theta}$, we consider the first-order derivative $\nabla_{\bm{x}} \bm{u}_{\bm{\theta}}$ and Laplacian $\Delta \bm{u}_{\bm{\theta}}$ involved in the loss function of a PINN training. Our implementation leverages the Stein  estimator~\cite{stein1981estimation}. Specifically, we represent the PDE solution $\bm{u}_{\bm{\theta}}(\bm{x})$ via a Gaussian smoothed model:
\begin{equation}
\bm{u}_{\bm{\theta}}(\bm{x})=\mathbb{E}_{\bm{\delta} \sim \mathcal{N}\left(\bm{0}, \sigma^2 \bm{I}\right)} f_{\bm{\theta}}(\bm{x}+\bm{\delta}),
\label{gaussian smoothed model}
\end{equation}
where $f_{\bm{\theta}}$ is a neural network with parameters $\bm{\theta}$; $\bm{\delta} \in \mathbb{R}^{D}$ is the random noise sampled from a multivariate Gaussian distribution ${\cal N}(\bm{0}, \sigma^2 \bm{I})$. With this special formulation, the first-order derivative and Laplacian of $\bm{u}_{\bm{\theta}}(\bm{x})$ can be reformulated as the expectation terms:
\begin{equation}
\begin{aligned}
 \nabla_{\bm{x}} \bm{u}_{\bm{\theta}}=\mathbb{E}_{\bm{\delta} \sim \mathcal{N}\left(\bm{0}, \sigma^2 \bm{I}\right)} & \left[\frac{\bm{\delta}}{2 \sigma^2}(f_{\bm{\theta}}(\bm{x}+\bm{\delta}) - f_{\bm{\theta}}(\bm{x}-\bm{\delta}))\right],\\
\Delta \bm{u}_{\bm{\theta}}=\mathbb{E}_{\bm{\delta} \sim \mathcal{N}\left[\bm{0}, \sigma^2 \bm{I}\right)}  & \left [f_{\bm{\theta}}(\bm{x}+\bm{\delta}) +
f_{\bm{\theta}}(x-\bm{\delta})-2 f_{\bm{\theta}}(\bm{x})\right]\\
& \times \frac{\|\bm{\delta}\|^2-\sigma^2 D}{2 \sigma^4}.  
\end{aligned}
\label{stein gradient estimator}
\end{equation}
In~\cite{he2023learning}, the above expectation is computed by evaluating $f_{\bm{\theta}}(\bm{x}+\bm{\delta})$ and $f_{\bm{\theta}}(\bm{x}-\bm{\delta})$ at a set of i.i.d. Monte Carlo samples of $\bm{\delta}$. Compared with finite difference~\cite{chiu2022can,xiang2022hybrid} which requires repeated calculations of gradients for each dimension, the Stein estimator can directly provide the vectorized gradients for all dimensions, allowing efficient parallel implementations. However, the Monte-Carlo Stein gradient estimator~\cite{he2023learning} needs a huge number of (e.g., $>10^3$) function queries even if variance reduction is utilized. Therefore, it is highly desirable to develop a more efficient BP-free method for evaluating the derivative terms in the loss function.

\subsection{Sparse-Grid Stein Gradient Estimator}


Now we leverage the sparse grid techniques~\cite{garcke2006sparse,gerstner1998numerical} to reduce the number of function queries in the Stein gradient estimator. Sparse grids have been extensively used in the uncertainty quantification~\cite{nobile2008sparse} of stochastic PDEs, but they have not been used in PINN training. 

To begin, we define a sequence of univariate quadrature rules $V=\left\{V_l: l \in \mathbb{N}\right\}$. Here $l$ denotes an accuracy level so that any polynomial function of order $\leq l$ can be exactly integrated with $V_l$. Each rule $V_l$ specifies $n_{l}$ nodes $N_{l}=\left\{\delta_{1},\dots,\delta_{n_{l}}\right\}$ and the corresponding weight function $w_l: N_{l} \rightarrow \mathbb{R}$. A univariate quadrature rule $V_{k}$ for a function $f$ of a random variable $\delta$, can be written as:
\begin{equation}
\int_{\mathbb{R}} f(\delta) p(\delta) \, d\delta \approx 
V_{k}[f]=\sum_{\delta_j \in N_{k}} w_{k}(\delta_j) f(\delta_j).
\end{equation}
Here $p(\delta)$ is the probability density function (PDF) of $\delta$. 

Next, we consider the multivariate integration of a function $f$ over a random vector $\bm{\delta}=(\delta^{1},\dots,\delta^{D})$. We denote the joint PDF of $\bm{\delta}$ as $p(\bm{\delta})=\prod_{m=1}^{D}p(\delta^{m})$ and define the $D$-variate quadrature rule with potentially different accuracy levels in each dimension indicated by the multi-index $\bm{l} = (l_1, l_2, ..., l_D) \in \mathbb{N}^{D}$. The Smolyak algorithm~\cite{gerstner1998numerical} can be used to construct sparse grids by combining full tensor-product grids of different accuracy levels and removing redundant points. Specifically, for any non-negative integer $q$, define $\mathbb{N}_{q}^{D} = \left\{\bm{l}\in \mathbb{N}^{D}: \sum_{m=1}^{D}l_{m} = D+q\right\}$ and $\mathbb{N}_{q}^{D} = \emptyset$ for $q<0$. The level-$k$ Smolyak rule $A_{D,k}$ for $D$-dim integration can be written as~\cite{wasilkowski1995explicit}:
\small
\begin{equation}
\begin{aligned}
A_{D,k}[f] =    \sum_{q=k-D}^{k-1} & (-1)^{k-1-q}\left(\begin{array}{c}
D-1 \\ k-1-q
\end{array}\right)  \times \\
& \sum_{\bm{l} \in \mathbb{N}_q^D}       \left(V_{l_{1}} \otimes \cdots \otimes V_{l_{D}} \right)[f] .
\end{aligned}
\end{equation}\normalsize
It follows that:
\begin{equation}
\begin{aligned}
A_{D,k}[f] = \sum_{q=k-D}^{k-1} \sum_{\bm{l} \in \mathbb{N}_q^D} \sum_{\delta^{1} \in N_{l_{1}}} \cdots \sum_{\delta^{D} \in N_{l_{D}}} (-1)^{k-1-q} \times\\
\left(\begin{array}{c}
D-1 \\ k-1-q
\end{array}\right)  \prod_{m=1}^{D} w_{l_{m}}(\delta^{m})f(\delta^{1},\dots,\delta^{D}), \nonumber
\end{aligned}
\end{equation}
which is a weighted sum of function evaluations $f(\bm{\delta})$ for $\bm{\delta} \in \bigcup_{q=k-D}^{k-1} \bigcup_{\mathbf{l} \in \mathbb{N}_q^D} \left(N_{l_{1}} \times \cdots \times N_{l_{D}}\right)$. The corresponding weight is $(-1)^{k-1-q}\left(\begin{array}{c}
D-1 \\ k-1-q
\end{array}\right) \prod_{m=1}^{D} w_{l_{m}}(\delta^{m})$. For the same $\bm{\delta}$ that appears multiple times for different combinations of values of $\bm{l}$, we only need to evaluate $f$ once and sum up the respective weights beforehand. The resulting level-$k$ sparse quadrature rule defines a set of $n_{L}$ nodes $S_{L}=\left\{\bm{\delta}_{1},\dots,\bm{\delta}_{n_{L}}\right\}$ and the corresponding weights $\left\{w_{1},\dots,w_{n_{L}}\right\}$. The $D$-dim integration can then be efficiently computed with the sparse grids as:
\begin{equation}
\int_{\mathbb{R}^{D}}f(\bm{\delta})p(\bm{\delta}) d \bm{\delta} \approx A_{D,k}[f] = \sum_{j=1}^{n_{L}} w_{j}f(\bm{\delta}_{j}).
\end{equation}
In practice, since the sparse grids and the weights do not depend on $f$, they can be pre-computed for the specific quadrature rule, dimension $D$, and accuracy level $k$. 

Finally, we implement the Stein gradient estimator in \eqref{stein gradient estimator} via the sparse-grid integration. Noting that $\bm{\delta} \sim \mathcal{N}\left(\bm{0}, \sigma^2 \bm{I}\right)$, we can use univariate Gaussian quadrature rules as basis to construct a level-$k$ sparse Gaussian quadrature rule $A_{D,k}^{*}$ for $D$-variate integration. Then the first-order derivative and Laplacian in \eqref{stein gradient estimator} is approximated as:
\small
\begin{equation}
\begin{aligned}
\nabla_{\bm{x}} \bm{u}_{\bm{\theta}}& \approx  \sum_{j=1}^{n_{L}^{*}}w_{j}^{*}  \left[\frac{\bm{\delta}_{j}^{*}}{2 \sigma^2}(f_{\bm{\theta}}(\bm{x}+\bm{\delta}_{j}^{*}) - f_{\bm{\theta}}(\bm{x}-\bm{\delta}_{j}^{*}))\right],  \\
\Delta \bm{u}_{\bm{\theta}}  & \approx \sum_{j=1}^{n_{L}^{*}}w_{j}^{*} \left(\frac{\|\bm{\delta}_{j}^{*}\|^2-\sigma^2 D}{2 \sigma^4}\right) \times \\
& \left( f_{\bm{\theta}}(\bm{x}+\bm{\delta}_{j}^{*}) +
f_{\bm{\theta}}(x-\bm{\delta}_{j}^{*})-2 f_{\bm{\theta}}(\bm{x}) \right),
\end{aligned}
\label{sparse-grid stein gradient estimator}
\end{equation} \normalsize
where the node $\bm{\delta}_{j}^{*}$ and weight $w_{j}^{*}$ are defined by the sparse grid $A_{D,k}^{*}$. We remark that $n_{L}^{*}$ is usually significantly smaller than the number of Monte Carlo samples required to evaluate \eqref{stein gradient estimator} when $D$ is small. For example, we only need $n_{L}^{*}=2D^{2}+2D+1$ nodes to approximate a $D$-dimensional integral ($D>1$) using a level-$3$ sparse Gaussian quadrature rule $A_{D,3}^{*}$.


\section{Numerical Results}
We test our tensor-compressed ZO training method on the MNIST dataset and a PDE benchmark. We compare our method with various baselines including standard first-order training and state-of-the-art BP-free training methods.

\vspace{-5pt}
\subsection{MNIST Image Classification} \label{sec: MNIST exp}
    To evaluate our proposed tensor-train (TT)  compressed hybrid ZO training method on realistic image classification tasks, we train a  Multilayer Perceptron (MLP) network with two FC layers (768$\times$1024, 1024$\times$10) to classify the MNIST dataset.  We compare our TT-compressed hybrid ZO training with four baseline methods: 1) first-order (FO) training using BP for accurate gradient evaluation, 2) ZO-RGE, 3) ZO-signRGE, and 4) ZO-CGE. We use SGD to update model parameters. We adopt a mini-batch size 64, an initial learning learning rate $\alpha$ as 1e-3 with exponential decaying rate of 0.9 every 10 epochs. All experiments run for 100 epochs. 


    \begin{table}[t]
      \centering
      \resizebox{\linewidth}{!}{
        \begin{tabular}{cccccc}
        \toprule
        \toprule
        Network & Neurons & Params & FO    & ZO-signRGE & Hybrid ZO \\
        \midrule
        MLP    & 1024  & 814090 & 97.43 &    &    \\
        MLP    & 10    & 7940  &  93.69 & 61.51 & 86.47 \\
        \midrule
        TT-MLP    & 1024  & 3962  & 97.16 & 88.48 & \textbf{96.64} \\
        \midrule
        SP-MLP    & 1024  & 4089  & 86.36 & 83.39 & 92.24 \\
        \bottomrule
        \bottomrule
        \end{tabular}%
        }
      \caption{Validation accuracy of MLPs, TT-MLPs, and SP-MLPs. Empty fields indicate no convergence.}
      \label{tab:addlabel}%
      \vspace{-10pt}
    \end{table}%

    \paragraph{(1) Effectiveness of TT Variance Reduction.} The proposed TT variance reduction is the key to ensure the success of ZO training on realistic neural networks. For the 2-layer MLP model, we fold the sizes of the input, hidden, and output layers as size $7\times 4\times 4\times 7$, $8\times 4\times 4\times 8$, $1\times 5\times 2\times 1$, respectively. We preset the TT-ranks as [1,$r$,$r$,$r$,1]. By varying $r$ we can control the compression ratio. We term this model as TT-MLP. As shown in Table~\ref{tab:addlabel}, the TT-compressed approach can reduce the model parameters by $205\times$. Due to the huge parameter and variance reduction, a hybrid ZO training can achieve a high testing accuracy $96.64\%$, which is very close to the FO training results ($97.16\%$). We further compare our method with two other sets of baselines:
    \begin{itemize}[leftmargin=*]
        \item{\bf FO and ZO training on the MLP model.} As shown in Table~\ref{tab:addlabel}, the original MLP model has a high testing accuracy of $97.43\%$ when trained with a FO SGD algorithm. Note that this accuracy is very close to that of our TT-compressed hybrid ZO training method. The ZO training methods cannot converge to a descent solution due to the large variance caused by the 814K model parameters.  
        \item{\bf FO and ZO training with pruning.} An alternative method of reducing model parameters and ZO gradient variance is pruning. To train a sparse-pruned MLP (termed SP-MLP), we adopted the Gradient Signal Preserving (GraSP) method~\cite{wang2020picking} at the initialization, which targets preserving the training dynamics after pruning. To ensure a fair comparison, we constrained the total number of parameters to be at the same level of TT-MLP. Table~\ref{tab:addlabel} show that pruning can indeed improve the quality of ZO training, but it still under-performs our proposed TT-compressed hybrid ZO training by a $4.4\%$ accuracy drop.
    \end{itemize}
    We also investigate the impact of various compression ratios by changing $r$. In Table~\ref{tab:MNIST result}, a higher compression ratio leads to better validation accuracy in both ZO-RGE and ZO-signRGE due to the smaller gradient variance. However, compressing the model too much can lose model expressive power and testing accuracy in both FO training and hybrid ZO training, as shown in the bottom row of Table~\ref{tab:MNIST result}.

    \begin{table}[t]
      \centering
      \resizebox{\linewidth}{!}{
        \begin{tabular}{ccccccc}
        \toprule
        \toprule
        Network & Ratio & Params & FO-AD & ZO-RGE & ZO-signRGE & Hybrid ZO \\
        \midrule
        FC-1024   & 1     & 814090 & 97.43 &  &  &  \\
        \midrule
        TT-MLP    & 0.01  & 8314  & 97.33 & 85.14 & 87.87 & 96.77 \\
              & 0.005 & 3962  & 97.16 & 87.72 & 88.48 & 96.64 \\
              & 0.003 & 2698  & 96.47 & 89.67 & 89.01 & 95.83 \\
        \bottomrule
        \bottomrule
        \end{tabular}%
        }   
      \caption{Validation accuracy of TT-MLPs with different compression ratios. Empty fields indicate no convergence.}
      \label{tab:MNIST result}%
      \vspace{-10pt}
    \end{table}%

    \begin{figure}[t]
        \centerline{\includegraphics[width=0.48\textwidth]{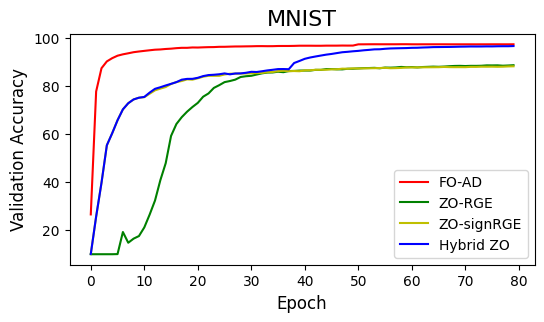}}
        \caption{Validation accuracy of TT-MLPs trained by different optimizers on MNIST dataset. We set $N=10$ and $\mu=0.1$ in ZO-RGE, ZO-signRGE, and the coarse-tuning stage in hybrid ZO. We set $\mu=0.01$ and momentum $m=0.9$ in the fine-tuning stage in Hybrid ZO. 
        }
        \label{MNIST_Acc}
        \vspace{-15pt}
    \end{figure}

     \paragraph{(2) Effectiveness of the Hybrid ZO Training.} Our hybrid ZO training was designed to achieve both good training accuracy and low function query complexity. To evaluate its effectiveness, we compare the hybrid ZO method with other baseline methods on the TT-MLP model with a compression ratio of $205\times$. 
    Figure \ref{MNIST_Acc} shows the validation accuracy from different ZO optimizers. ZO-signRGE shows a more stable training than ZO-RGE at the beginning. However, both ZO-RGE and ZO-signRGE fail to achieve high accuracy in the late stage compared to the FO SGD baseline due to gradient estimation variance. With additional ZO-CGE fine-tuning, our hybrid ZO training strategy achieves a highly competitive result, comparable to the FO baseline. This hybrid ZO training optimizer balances the loss computation efficiency of ZO-signRGE and the accurate estimation capability of ZO-CGE. Specifically, to reach 85\% accuracy in the early stage ZO-CGE needs 11 million forward evaluations while ZO-signRGE only needs 340K (32.75$\times$ fewer) forward evaluations. Table \ref{tab:MNIST result} presents the converged results of TT MLPs with different compression ratios. The advantage of the hybrid ZO over ZO-RGE and ZO-signRGE is consistent regardless of the compression ratio. 

    \begin{table}[t]
      \centering
       \resizebox{\linewidth}{!}{
        \begin{tabular}{cccc}
        \toprule
        \toprule
        Method & Params & \# of loss eval. & Best Acc. \\
        \midrule
        \multicolumn{4}{c}{Hybrid ZO Optimizer} \\
        \midrule
        HZO & 814K  & 814K  & 83.58 \\
        TT-HZO (proposed)& \textbf{4k} & 4K    & \textbf{96.64} \\
        SP-HZO & \textbf{4k} & 4K    & 92.24 \\
        \midrule
        \multicolumn{4}{c}{Sparse ZO Optimizer} \\
        \midrule
        FLOPS & 814K  & \textbf{0.1K} & 83.50 \\
        STP  & 814K  & 24.4K & 90.20* \\
        SZO-SCD & 814K  & 18.3K & 93.50* \\
        \midrule
        \multicolumn{4}{c}{Forward Gradient} \\
        \midrule
        FG-W  & 272K  & /     & 90.75 \\
        LG-FG-A & 272K  & /     & 96.76 \\
        FG-W  & 429K  & /     & 91.44 \\
        LG-FG-A & 429K  & /     & \textbf{97.45} \\
        \midrule
        \multicolumn{4}{c}{Feedback Alignment} \\
        \midrule
        FA   & 784K   & /     & 97.90 \\
        DFA  & 1.26M  & /     & \textbf{98.30} \\
        \bottomrule
        \bottomrule
        \end{tabular}%
          }  
      \caption{Comparison of BP-free optimizers on the MNIST dataset. * means fine-tuning on a pre-trained model. In the Forward Gradient group, FG refers to forward gradient, LG refers to local objective functions added, -W refers to weight-perturbed, and -A refers to activity-perturbed.
      }
      \label{tab:MNIST compare}%
      \vspace{-10pt}
    \end{table}%
    
    \paragraph{(3) Comparison with other BP-free Training.} We further compare our TT-compressed hybrid ZO optimizer with three families of BP-free algorithms:
    \begin{itemize}[leftmargin=*]
        \item {\bf Other ZO Optimizers:} These methods use ZO optimization to estimate the gradients, and update model parameters with SGD iterations. Due to the dimension-dependent estimation error and loss computation complexity, most methods explore sparsity in training. FLOPS~\cite{GuDAC}, STP~\cite{bibi2020stochastic} and SZO-SCD~\cite{GuAAAI} assume a random pre-specified sparsity. We also extend the sparse method to ZO training.
        \item{\bf Forward Gradient (FG):} Standard FG method uses forward-mode AD to evaluate the \textit{forward gradient} and performs SGD-type iterations \cite{baydin2022gradients}. Ref. \cite{ren2022scaling} further scales up this method by introducing activation perturbation and local loss to reduce the variance of forward gradient evaluation. 

        \item {\bf Feedback Alignment (FA):} Standard FA~\cite{lillicrap2016random} uses random and fixed backward weights. Direct feedback alignment (DFA) propagated the error through fixed random feedback connections directly from the output layer to each hidden layer.~\cite{nokland2016direct}.
    \end{itemize}

    The results are summarized in Table \ref{tab:MNIST compare}. For ZO optimizers, we also compare the number of loss computations needed for each iteration. Our method can achieve the best accuracy in weight perturbation-based methods. Compared with SOTA FG and FA algorithms, our method can also achieve comparable accuracy in the MNIST dataset. Note that our method contains over 107× fewer model parameters. This can greatly save the memory overhead for on-device training scenarios with restricted memory resources. 
    
    We remark that both FG and FA are not exactly black-box optimizers, since they rely on the computational graph of a given deep learning software. In practice, storing the computational graphs on an edge device can be expensive.

\vspace{-5pt}
\subsection{PINN for Solving High-Dim HJB PDE}
We further use our BP-free method to train a PINN arising from high-dim optimal control of robots and autonomous systems. We consider the following 20-dim HJB PDE:
       \begin{equation}
        \begin{aligned}
        &\partial_t u(\bm{x}, t)+\Delta u(\bm{x}, t)-0.05 \left\|\nabla_{\bm{x}}u(\bm{x}, t)\right\|_{2}^{2}=-2, \\
        &u(\bm{x}, 1)=\left\|\bm{x}\right\|_{1}, \quad \bm{x} \in [0,1]^{20}, ~~t \in[0, 1].
        \end{aligned}
        \end{equation}
        Here $\left\|\cdot\right\|_{p}$ denotes an $\ell_p$ norm. The exact solution is $u(\bm{x},t)=\left\|\bm{x}\right\|_{1}+1-t$. The baseline neural network is a 3-layer \textbf{MLP} ($21\times n, n\times n, n\times 1$, $n$ denotes the number of neurons in the hidden layer) with sine activation. We consider four options for computing derivatives in the loss:  1) finite-difference (\textbf{FD})~\cite{lim2022physics}, 2) Monte Carlo-based Stein Estimator (\textbf{SE})~\cite{he2023learning}, 3) our sparse-grid (\textbf{SG}) method, and 4) automatic differentiation (\textbf{AD}) as a golden reference. We approximate the solution $u_{\bm{\theta}}$ by a transformed neural network $f_{\bm{\theta}}^{'}(\bm{x},t) = f_{\bm{\theta}}(\bm{x},t) + \left\|\bm{x}\right\|_{1} + 1 -t$, where $f_{\bm{\theta}}(\bm{x},t)$ is the base neural network or its TT-compressed version. Specifically, $u_{\bm{\theta}}(\bm{x},t)=f_{\bm{\theta}}^{'}(\bm{x},t)$ for \textbf{FD} and \textbf{AD}, and $u_{\bm{\theta}}(\bm{x},t)=\mathbb{E}_{(\bm{\delta_{\bm{x}}},\delta_{t}) \sim \mathcal{N}\left(\bm{0}, \sigma^{2} \bm{I}\right)}f_{\bm{\theta}}^{'}(\bm{x}+\bm{\delta_{\bm{x}}},t+\delta_{t})$ for \textbf{SE} and \textbf{SG}. Here the transformed network is designed to ensure that our approximated solution either exactly satisfies (\textbf{FD}, \textbf{AD}) or closely adheres to the terminal condition (\textbf{SE}, \textbf{SG}), allowing us to focus solely on minimizing the HJB residual during training.
        
       In our hybrid ZO optimization, we set $\mu=0.1$ and $N=10$ in the ZO-signRGE coarse training stage. With greatly reduced variance by TT, ZO-signRGE can provide sufficiently accurate results, so fine-tuning is not necessary for this example. In the loss evaluation, we set the step size in \textbf{FD} to 0.01 and the noise level $\sigma$ to 0.1 in \textbf{SE} and \textbf{SG}. We use 1024 samples in \textbf{SE} and 925 samples in \textbf{SG} using a level-3 sparse Gaussian quadrature rule to approximate the expectations \eqref{gaussian smoothed model} and \eqref{stein gradient estimator}. Note that for most 2-dim or 3-dim PDEs, our SG method only requires  $10 \sim 30$ samples. Our results are summarized below:



        \begin{itemize}[leftmargin=*]
            \item {\bf Effectiveness of BP-free loss computation:} 
        We compare our sparse-grid loss computation with those using AD, FD, and Monte Carlo-based Stein estimator, respectively. To evaluate different loss evaluations, we perform FO training and report the results in Table \ref{tab:HJB FO}. The BP-free loss computation does not hurt the model performance, and our SG method is competitive compared to the original PINN training using AD for derivative computation.

        \item{\bf Evaluation of the BP-free PINN Training.}  Table \ref{tab:HJB ZO} shows the MSE error of fully BP-free PINN training, using various BP-free loss computation and the proposed hybrid ZO training. By employing a tensor-train (TT) compressed model to reduce the variance, our method achieved a validation loss similar to standard FO training using AD. This observation clearly demonstrates that our method can bypass BP in both loss evaluation and model parameter updates with little accuracy loss.

        \item{\bf Comparison of dimensionality/variance reduction approaches.} We compare our TT-compressed model with standard MLPs with dense fully connected (FC) layers and sparse-pruned (SP) MLPs in Table~\ref{tab:HJB dim reduction}. For a fair comparison, all networks have similar model parameters. TT-compressed and SP model can reach better results than FC models as TT and SP can preserve the wide hidden layer. The SP model shows superior efficiency in FO training and achieves similar result with the TT model in ZO training. Note that the SP model only reduces the trainable parameters but not the model parameters, so TT models can achieve better memory and computation saving.
        \end{itemize}
        
        \begin{table}[t]
          \centering
          \resizebox{\linewidth}{!}{
            \begin{tabular}{ccccccc}
            \toprule
            \toprule
            Network & Params & FO+AD & FO+FD & FO+SE & FO+SG \\
             & & & & &                              (proposed) \\
            \midrule
            MLP       & 608257 & 3.58E-04 & 5.97E-05 & 8.98E-04 & 4.27E-04 \\
            \midrule
            TT-MLP    & 7745  & 4.28E-05 & \textbf{4.90E-05} & 2.28E-04 & 7.14E-05 \\
                     & 3209  & 1.63E-05 & 4.56E-05 & 3.82E-04 & \textbf{4.49E-05} \\
            \bottomrule
            \bottomrule
            \end{tabular}%
        }
        \caption{FO training with different loss evaluations.}
          \label{tab:HJB FO}%
        \end{table}%

        \begin{table}[t]
          \centering
          \resizebox{\linewidth}{!}{
            \begin{tabular}{ccccccc}
            \toprule
            \toprule
            Network & Params & FO+AD & ZO+FD & ZO+SE & ZO+SG \\
             & & & & &                              (proposed) \\
            \midrule
            MLP    & 608257 & 3.58E-04 & 9.23E-02 & diverge & 1.62E-01 \\
            \midrule
            TT-MLP & 7745  & 4.28E-05 & 3.79E-04 & 4.35E-04 & \textbf{4.79E-05} \\
                   & 3209  & 1.63E-05 & 3.82E-04 & 2.75E-04 & \textbf{1.44E-05} \\
            \bottomrule
            \bottomrule
            \end{tabular}%
            }   
          \caption{ZO training results for 20-dim HJB PDE.}
          \label{tab:HJB ZO}%
          \vspace{-10pt}
        \end{table}%

        \begin{table}[t]
          \centering
            \begin{tabular}{ccccc}
            \toprule
            \toprule
            Network & Neurons & Params & FO-AD & ZO-SG \\
            \midrule
            FC-3  & 768   & 608257& 3.58E-04 & 1.62E-01 \\
            \midrule
            FC-3  & 78    & 7800  & 1.52E-02 & 4.35E-02 \\
            FC-2  & 320   & 7700  & 6.55E-04 & 1.00E-02 \\
            TT-3  & 768   & 7745  & 4.25E-05 & \textbf{4.79E-05} \\
            SP-3  & 768   & 7603  & \textbf{1.54E-09} & 6.29E-05 \\
            \bottomrule
            \bottomrule
            \end{tabular}%
          \caption{Validation loss of various models for the HJB PDE}
          \label{tab:HJB dim reduction}%
          \vspace{-10pt}
        \end{table}%
        
        
\section{Relevant Work}
    \paragraph{Training on edge devices.}
   On-device inference has been well studied. 
   However, training requires extra computation for BP and extra memory for intermediate results. Edge devices usually have tight memory and power budget, and run without an operating system, making it infeasible to implement modern deep learning frameworks that support automatic differentiation~\cite{baydin2018automatic}.
    Most on-device training frameworks implemented BP by hand and only update within a small subspace (e.g., the last classifier layer~\cite{donahue2014decaf}, biases~\cite{zaken2021bitfit}, etc.), only capable of incremental learning or transfer learning by fine-tuning a pre-trained model.
    Some work leveraged pruning or sparse training to reduce the trainable parameter number\cite{lin2022device, gu2021l2ight}, but cannot achieve real memory savings.
    Recent work found tiny, trainable subspace exists in deep neural networks~\cite{frankle2018lottery, li2022low}, but a pre-training is needed to identify such a subspace, which does not save the training overhead. Tensorized training ~\cite{chen20213u, zhang2021fpga} offers orders-of-magnitude memory reduction in the training process, enabling end-to-end neural network training on FPGA. However, it is non-trivial to implement this FO method to train larger models on edge devices due to the complex gradient computation.

    \paragraph{Back-propagation-free training}
    Due to the complexity of performing BP on edge devices, several BP-free training algorithms have been proposed. These methods have gained more attentions in recent years as BP is also considered “biologically implausible”. ZO optimization \cite{ZOSGD, duchi2015optimal, ZOSCD, chen2019zo,shamir2017optimal,balasubramanian2022zeroth,yang2020quantile,cai2021zeroth,yang4040704vflh} plays an important role in tackling signal processing and machine learning problems, where actual gradient information is infeasible.
    For specific use cases, please refer to the  survey by \citet{liu2020primer}.
    ZO optimizations were also applied in on-device training for optical neural networks (ONN) \cite{GuDAC, GuAAAI, gu2021l2ight}. A ZO SGD-based method FLOPS \cite{GuDAC} was proposed for end-to-end ONN training for vowel recognition, and a stochastic ZO sparse coordinate descent SZO-SCD \cite{GuAAAI} was further proposed for efficient fine-tuning for ONN.
    But most ZO optimizations scale poorly when training real-size neural networks from scratch, due to their dimension-dependent gradient errors. Forward-forward algorithm was proposed for biologically plausible learning~\cite{hinton2022forward}. Other BP-free training frameworks include the forward gradient method~\cite{baydin2022gradients}, which updates the weights based on the directional gradient computed by forward-mode AD along a random direction. This method is further scaled up by leveraging activity perturbation and local loss for variance reduction\cite{ren2022scaling}. 
    However, existing BP-free methods focus on training on HPC, lacking special concerns for the special challenges of on-device training. Furthermore, BP-free training of PINN remains an empty field. 

\section{Conclusion}
Due to the high complexity and memory overhead of back propagations (BP) on edge devices, this paper has proposed a completely BP-free zeroth-order (ZO) approach to train real-size neural networks from scratch. Our method utilizes tensor compression to reduce the ZO gradient variance and a hybrid approach to improve the efficiency of ZO training. This has enabled highly accurate ZO training on the MNIST dataset, showing superior performance than sparse ZO approach. We have further extended this memory-efficient approach to train physics-informed neural networks (PINN) by developing a sparse-grid Stein gradient estimator for the loss evaluation. Our approach has been successfully solved a 20-dim HJB PDE, achieving similar accuracy with standard training using first-order optimization and automatic differentiation. Due to the huge memory reduction and the BP-free nature, our method can be easily implemented on various resource-constraint edge devices in the near future. 

The ZO-CGE fine-tuning stage of our method requires many more function queries than the ZO-signRGE coarse-training stage. This prevents its applications in training large model (transformers) from scratch. We will address this issue in the future.

\newpage
\bibliography{aaai24}

\end{document}